\documentclass[10pt,twocolumn,letterpaper]{article}

\usepackage{multirow}
\usepackage{bm}
\usepackage[pagenumbers]{cvpr} 
\usepackage{cuted}
\usepackage[dvipsnames]{xcolor}

\usepackage{amsmath,amsfonts,bm}

\def\eqref#1{equation~\ref{#1}}

\def\1{\bm{1}}

\def\vd{{\bm{d}}}

\def\vx{{\bm{x}}}

\def\vz{{\bm{z}}}

\DeclareMathAlphabet{\mathsfit}{\encodingdefault}{\sfdefault}{m}{sl}
\SetMathAlphabet{\mathsfit}{bold}{\encodingdefault}{\sfdefault}{bx}{n}

\def \transpose {\mathrm{T}}

\def \path {\mathit{path}}

\definecolor{cvprblue}{rgb}{0.21,0.49,0.74}
\usepackage[pagebackref,breaklinks,colorlinks,citecolor=cvprblue]{hyperref}

\def\paperID{} 
\def\confName{}
\def\confYear{}

\title{World-consistent Video Diffusion with Explicit 3D Modeling}

\newcommand\NAME{\texttt{WVD}\xspace}
\newcommand\Model{WVD\xspace}
\newcommand\XYZ{XYZ}

\author{Qihang Zhang$^{1,2}$\thanks{Work done during internship at Apple MLR.} \quad Shuangfei Zhai$^{1}$ \quad Miguel Angel Bautista Martin$^{1}$ \quad Kevin Miao$^{1}$\\ 
\quad Alexander Toshev$^{1}$ \quad Josh Susskind$^{1}$ \quad Jiatao Gu$^{1}$ \\
	{$^1$Apple  \;
	$^2$The Chinese University of Hong Kong 
}
    \\
	}

\begin{document}
\maketitle
\begin{strip}
    \centering
    \vspace{-5em}
    \centering
    \includegraphics[width=\textwidth]{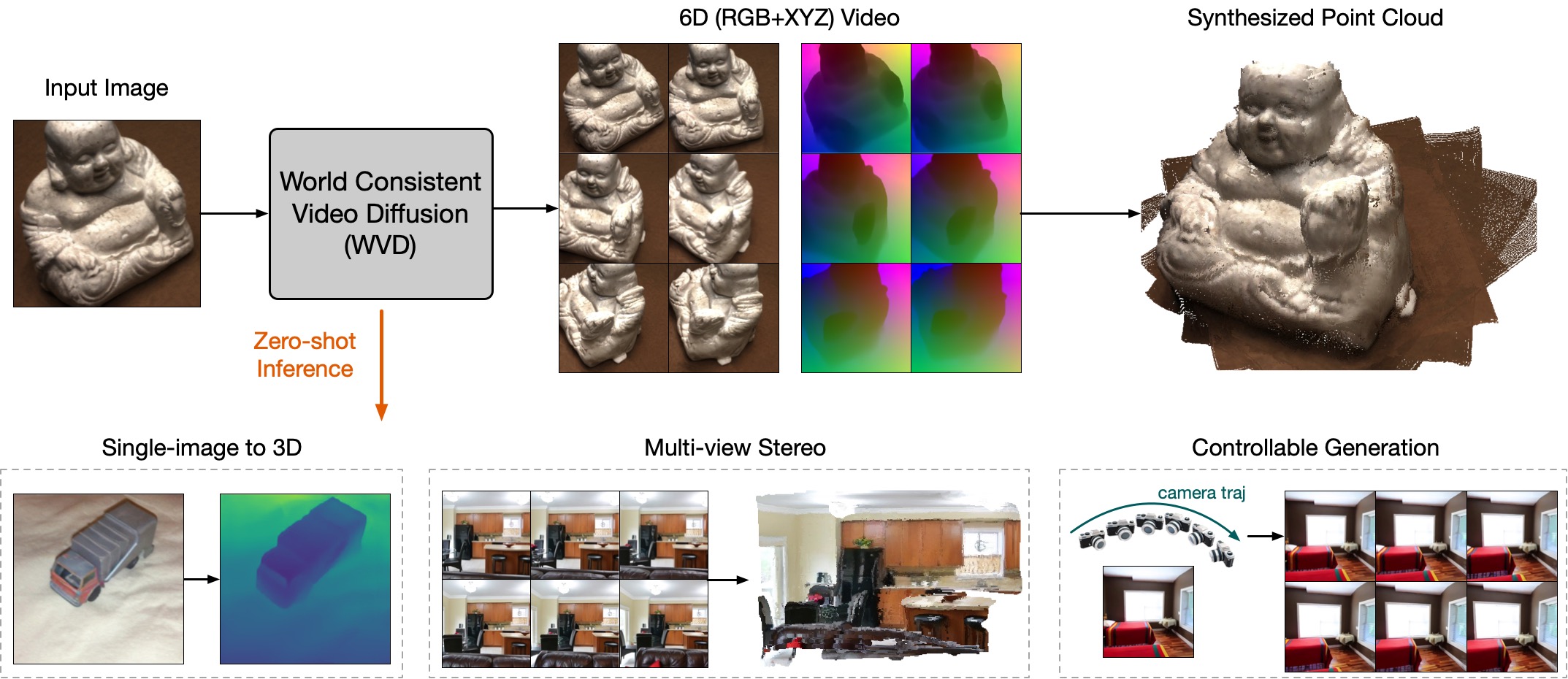}
    \vspace{-2em}
    \captionof{figure}{\NAME predicts 6D videos from an image, unifying various 3D tasks with a single diffusion model.} 
    \label{fig:teaser}
\end{strip}
\begin{abstract}
Recent advancements in diffusion models have set new benchmarks in image and video generation, enabling realistic visual synthesis across single- and multi-frame contexts. However, these models still struggle with efficiently and explicitly generating \textit{3D-consistent} content. To address this, we propose \textbf{World-consistent Video Diffusion} (\Model), a novel framework that incorporates explicit 3D supervision using {\XYZ} images, which encode global 3D coordinates for each image pixel.
More specifically, we train a diffusion transformer to learn the joint distribution of RGB and {\XYZ} frames. This approach supports multi-task adaptability via a flexible inpainting strategy. For example, \Model can estimate {\XYZ} frames from ground-truth RGB or generate novel RGB frames using {\XYZ} projections along a specified camera trajectory. In doing so, \Model unifies tasks like single-image-to-3D generation, multi-view stereo, and camera-controlled video generation.
Our approach demonstrates competitive performance across multiple benchmarks, providing a scalable solution for 3D-consistent video and image generation with a single pretrained model.
Our project website is at \url{https://zqh0253.github.io/wvd}.

\end{abstract}    
\section{Introduction}

Recent advancements in deep generative models have brought significant breakthroughs to the field of visual synthesis, with diffusion models~\cite{sohl2015deep, ho2020denoising} emerging as the state-of-the-art approach for high-quality image generation~\cite{rombach2022high,  esser2024scaling}. These models have demonstrated remarkable success in generating realistic images. By extending the input from single-frame image to multiple frames, diffusion models have also been applied to tasks such as video generation~\cite{ho2022video, guo2023animatediff, wang2023lavie, ho2022imagen, blattmann2023stable, girdhar2023emu, bar2024lumiere, brooks2024video} and multi-view image synthesis~\cite{liu2023zero1to3, gao2024cat3d, shi2023mvdream, wang2023imagedream, yang2024consistnet, liu2023syncdreamer, hollein2024viewdiff}, where consistency across frames is crucial. In these applications, multi-frame consistency is typically learned in an implicit manner, e.g., through the attention mechanism that captures relationships across frames. Despite their success, multi-view (video) diffusion models face several limitations: they demand large amounts of data and significant computational resources for training, and they lack explicit guarantees for 3D consistency, often leading to 3D inconsistencies.

In contrast to these implicit methods, some approaches~\cite{eg3d, giraffe, gu2023nerfdiff, xu20223d, gnerf} seek to explicitly model 3D correspondences by embedding 3D inductive biases into the generative pipeline. These methods leverage techniques such as volume rendering~\cite{mildenhall2020nerf}, which can impose constraints that ensure 3D consistency in the generated images. However, the integration of 3D inductive biases tends to place heavy constraints on both the data and the architectural design, making it difficult to scale these methods to more complex datasets with diverse distributions.

To address these limitations, we propose a novel framework for multi-view and video generation that introduces explicit 3D supervision into diffusion models. Our method is designed to handle both RGB image generation and 3D geometry modeling within a unified framework. A major challenge in this integration arises from the inherent incompatibility between traditional 3D geometry representations and existing image architectures, such as the 2D Transformer-based models commonly used in diffusion models (e.g., DiT~\cite{peebles2023scalable}). To resolve this, we propose to use {\XYZ} images to represent 3D geometry, which are compatible with 2D Transformer architectures. Each pixel in an {\XYZ} image records its corresponding global 3D coordinates. Unlike RGB images, which encode complex texture and lighting information, {\XYZ} images are textureless and only capture geometric information, making them ideal for providing explicit 3D supervision during training.

Furthermore, because our model learns the joint distribution of RGB and {\XYZ} images during the training phase, it can naturally perform conditional generation during inference using a flexible inpainting strategy~\cite{meng2021sdedit, lugmayr2022repaint}. This enables the model to adapt to a wide range of tasks beyond image synthesis, including camera pose estimation, single-view and multi-view depth prediction from unposed images, and camera-conditioned novel view synthesis. This versatility allows our model to unify various generative and discriminative tasks under a single framework.

We refer to our proposed method as \Model. The major contributions of our work can be summarized as follows:
\begin{itemize}
    \item  We propose a novel approach to learn a multi-view diffusion model with explicit 3D supervision.

    \item Via a flexible inference strategy, \Model is capable of unifying various tasks within a single framework.

    \item \Model achieves competitive performance over different tasks, showcasing the potential to become a world-consistent 3D foundation model.
\end{itemize}

\section{Related Work}

\noindent\textbf{Multi-view Diffusion Models.}
The advancement of multi-view diffusion models represents a significant step in generative modeling, combining the robust generation capabilities of diffusion frameworks with the complex requirement for cross-view consistency. Notable approaches like MVDream~\cite{shi2023mvdream}, ImageDream~\cite{wang2023imagedream}, Zero123++~\cite{liu2023zero1to3}, ConsistNet~\cite{yang2024consistnet}, SyncDreamer~\cite{liu2023syncdreamer}, and ViewDiff~\cite{hollein2024viewdiff} adapt text-to-image diffusion models~\cite{rombach2022high} to produce synchronized multi-view outputs. Video diffusion models~\cite{ho2022video, guo2023animatediff, wang2023lavie, ho2022imagen, blattmann2023stable, girdhar2023emu, bar2024lumiere, brooks2024video} learn multi-view consistency from extensive video datasets.  Models like CameraCtrl~\cite{he2024cameractrl}, MotionCtrl~\cite{wang2024motionctrl}, and Camco~\cite{xu2024camco} enhance video diffusion models by introducing camera-specific conditions, which allow for controlled synthesis of novel views across different perspectives. 

\noindent\textbf{Estimating 3D from Multi-view Images.}
Estimating 3D structure from multi-view images remains a foundational challenge in 3D vision. Classical approaches, such as COLMAP~\cite{schoenberger2016mvs}, tackle this problem with a multi-stage pipeline involving keypoint detection and matching, RANSAC~\cite{fischler1981random}, Perspective-n-Point (PnP) solvers ~\cite{fischler1981random}, and a final bundled adjustment step for refinement. 
While classical geometric methods are effective, they require extensive engineering and optimization, often making it challenging to achieve accurate solutions, especially with large or complex datasets. Modern approaches study end-to-end learning methods that simplify the 3D estimation pipeline while also learning 3D priors from data. For example, VGGsfm~\cite{wang2024vggsfm} introduces differentiability at every stage of the COLMAP pipeline, making the process more adaptable to gradient-based optimization. DUSt3R~\cite{wang2024DUSt3R} takes this a step further by employing Vision Transformers to regress point clouds directly from unposed image pairs. Mast3R~\cite{mast3r_arxiv24} builds on these methods, enhancing performance by predicting features that increase the accuracy of keypoint matching, resulting in more reliable 3D reconstructions.  These recent end-to-end approaches reduce the need for complex engineering and iterative processes, offering an efficient alternative to traditional multi-view 3D reconstruction pipelines and paving the way for more robust and scalable 3D vision applications.

\section{World-consistent Video Diffusion (\Model)}

\begin{figure*}[t]
    \centering
    \includegraphics[width=\textwidth]{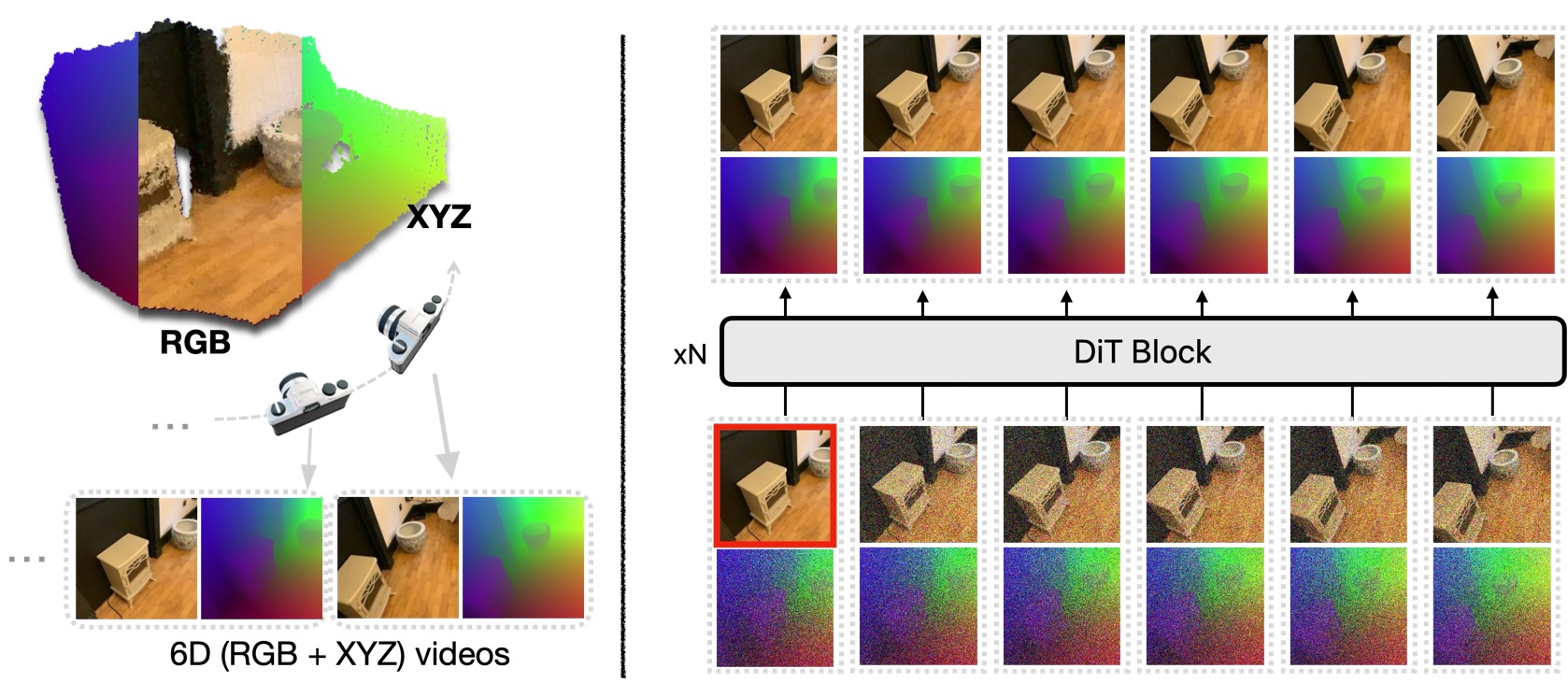}
    \caption{An illustration of \textbf{\NAME} pipeline. The left part shows 6D videos formed by RGB and XYZ frames. On the right part, WVD iteratively denoises the 6D videos based on a specified RGB frame, which is highlighted with a red box.}
    \label{fig:pipeline}
\end{figure*}

In this section, we present World-consistent Video Diffusion Models (\Model), which leverage diffusion models to jointly model the distribution of RGB and {\XYZ} frames across different viewpoints. We begin by introducing foundational concepts of diffusion models and its application in modeling 3D content (\cref{sec:preliminaries}), followed by an in-depth discussion of our architectural design (\cref{sec:arch}). 

\subsection{Preliminaries}
\label{sec:preliminaries}

\paragraph{Diffusion Models.} Standard diffusion models~\cite{ho2020denoising} operate by iteratively transforming noise into structured data through a denoising process. More specifically, a data point \( \mathbf{x}_0 \) is progressively noised through a forward process, yielding a sequence \( \{ \mathbf{x}_t \}_{t=1}^T \) according to a variance-scheduled Gaussian distribution. The diffusion model aims to reverse this, parameterized as \( p_\theta(\mathbf{x}_{t-1} | \mathbf{x}_t) \), where \( \theta \) denotes the model parameters. To reduce the computation cost for high-resolution inputs, LDM~\cite{rombach2022high} improves by learning diffusion in the latent space $\vz = \mathcal{E}(\vx)$ of a pretrained VAE~\citep{kingma2013auto}.
Diffusion models are typically implemented using a UNet architecture~\cite{ronneberger2015u}. Recently, however, Diffusion Transformers (DiT)\cite{peebles2023scalable} have emerged as a promising alternative. Leveraging the self-attention mechanism of Transformers to model intricate dependencies, DiT has demonstrated significant improvements in the fidelity of generated outputs and enhanced flexibility. This approach has shown potential across modalities, including images~\cite{chen2023pixart} and videos~\cite{videoworldsimulators2024}. For instance, DiT can process videos by flattening and concatenating each frame into a single long sequence, allowing it to jointly denoise all frames.

\vspace{-10pt}\paragraph{Multi-view Diffusion Models.} 
A common approach for diffusion models to learn 3D structure involves modeling the joint distribution of multi-view images~\citep{shi2023mvdream,liu2023syncdreamer} and reconstructing 3D content in a second stage. This reconstruction is typically achieved either through optimization~\cite{mildenhall2020nerf} or feed-forward prediction~\cite{hong2023lrm}. One can use DiT to process multi-view inputs similar to video diffusion, where DiT’s attention layers operate across views. This implicitly captures 3D consistency, thus ensuring coherent image synthesis across perspectives. To make the diffusion process  3D controllable, approaches like CAT3D~\cite{gao2024cat3d} condition the model on camera ray maps ($\mathbf{r}$)~\cite{Sitzmann2021LightRendering} using  $p_\theta(\mathbf{x}_{t-1} | \mathbf{x}_t, \mathbf{r})$. This condition is crucial as it allows the trained model to generate novel views during inference.

However, this approach has two clear challenges: (1) It lacks explicit 3D guarantees, relying on the model to infer consistency purely from multi-view images. This often requires significant computational resources and high-quality data, yet it can still suffer from 3D inconsistency failures; (2) Furthermore, the dependence on camera ray inputs poses challenges for scaling to large datasets due to fundamental ambiguities in existing camera representations. These representations struggle to handle variations across datasets, necessitating non-trivial camera normalization~\cite{watson2024controlling}, which further complicates the training process.

\subsection{Approach}
\label{sec:arch}

To tackle the primary challenges faced by traditional multi-view diffusion models, we propose \textbf{W}orld-consistent \textbf{V}ideo \textbf{D}iffusion (\Model), drawing inspiration from advancements in video diffusion models. Instead of incorporating additional camera control, our approach explicitly predicts 3D geometry by simultaneously diffusing over RGB frames and their corresponding point clouds. Specifically, the point clouds are projected into each frame as {\XYZ} images. 

\vspace{-10pt}\paragraph{{\XYZ} Image Representation.}
Point clouds are a widely used representation of 3D geometry. However, their highly unstructured nature ($X \in \mathbb{R}^{N \times 3}$) poses significant challenges for learning with standard DiT architectures. To address this, we propose representing a 3D scene using multiple {\XYZ} images, which provide a structured and learnable format. The transformation from a point cloud to an XYZ image is defined as:
\begin{equation}
    \vx^\textrm{XYZ} = \mathcal{R}(\mathcal{N}(X), X, C), 
    \label{eq.raster}
\end{equation}
where $C=(P, K)$  represents the camera parameters, including the pose ($P$) and intrinsic ($K$) matrices. Here,  $\mathcal{N}$  is a normalization function that centers and rescales the point clouds within the range [-1, 1], and  $\mathcal{R}$  is a rasterizer that maps the normalized 3D point values onto an image plane, using the 3D positions and camera transformations. This representation ensures compatibility with existing architectures while preserving the geometric structure of the scene.

The {\XYZ} image has the same shape as its RGB counterpart, with each pixel corresponding to a 3D point in the global coordinate system. By combining {\XYZ} and RGB images into a unified 6D video representation, the model effectively captures a 3D region while maintaining compatibility with standard video diffusion architectures. Beyond its simplicity and learnability, representing 3D geometry using {\XYZ} images offers several additional benefits:

\begin{itemize}
    \item \textbf{Explicit Consistency Supervision:} {\XYZ} images are texture-free and provide robust pixel alignment across views, unlike RGB images, which are influenced by variations in texture and lighting. When two pixels in different views share the same value in {\XYZ} images, they correspond to the same location in the global 3D coordinate system. This property facilitates strong pixel correspondence across views, enabling direct 3D supervision during the generation of both {\XYZ} and RGB images.
    \item \textbf{Elimination of Camera Control:} By encoding 3D geometry directly, {\XYZ} images obviate the need for additional camera information to align multiple views, as required in existing methods~\cite{gao2024cat3d,he2024cameractrl}. This approach reduces camera-related ambiguities, making it practical to scale up to larger and more complex datasets.
\end{itemize}

\vspace{-10pt}\paragraph{RGB-{\XYZ} Diffusion.} 
Following prior works~\cite{gao2024cat3d}, \Model learns a DiT-like model in the latent space, operating on a sequence of 6D video data \(\{\vx_n^\textrm{RGB}, \vx_n^\textrm{\XYZ}\}_{n=1}^N\). Since \(\vx_n^\textrm{\XYZ}\) is pre-normalized, it can be directly processed using pretrained VAEs~\citep{rombach2022high} without requiring additional fine-tuning. To improve computational efficiency, we concatenate the RGB and {\XYZ} latents along the channel dimension before adding noise:
\begin{equation}
    \vz_n = [\mathcal{E}(\vx^\textrm{RGB}_n); \mathcal{E}(\vx^\textrm{\XYZ}_n)] \in \mathbb{R}^{L\times 2D},
\end{equation}
where $L$ is the sequence length, $D$ is the latent dimension. This design allows us to directly fine-tune pretrained image or video diffusion models, significantly enhancing training efficiency. The model can be trained using either text- or image-conditioned data, depending on the dataset. For instance, as illustrated in \cref{fig:pipeline}, in the case of image-conditioned generation, the added noise on the conditional image is simply removed at each iteration during training.

\vspace{-10pt}\paragraph{Post Optimization.} As \Model directly predicts the point clouds in the global coordinates, we can easily perform a  Perspective-n-Point (PnP) algorithm to recover the corresponding camera $C=(P, K)$ and depth maps $\vd$ given the predicted {\XYZ} images $\hat{\vx}^\textrm{\XYZ}$. In this paper, we directly perform gradient optimization over re-projection loss:
\begin{equation}
    \min_{P, K, \vd} \sum_{u, v}\|
        \tilde{\vx}^\textrm{\XYZ}_{u, v} - \hat{\vx}^\textrm{\XYZ}_{u, v}
    \|^2_2,
    \label{eq.post}
\end{equation}
where $\tilde{\vx}^\textrm{\XYZ}_{u, v} = P^{-1}K^{-1}\vd_{u, v}[u, v, 1]^\transpose$, and
$u, v$ are the pixel coordinates. This post-optimization step is efficient and can be easily parallelized across views. Moreover, the optimized depth map \(\vd\) and camera parameters $C$ provide a more accurate and physically consistent estimation $\tilde{\vx}^\textrm{\XYZ}$ of the original {\XYZ} image, which can be highly beneficial for downstream tasks.

\section{\Model as a 3D Foundation Model}
\label{sec:flexible-inference}
\Model learns to generate RGB and {\XYZ} frames together by modeling the joint probability \( P(\text{RGB}, \text{\XYZ}) \), effectively capturing their interdependent structures and features. At inference time, this joint distribution can be leveraged to estimate conditional distributions, such as \( P(\text{\XYZ} \mid \text{RGB}) \) or \( P(\text{RGB} \mid \text{\XYZ}) \). This capability makes \Model a foundation for supporting a wide range of downstream tasks.

\begin{figure*}[t]
    \centering
    \includegraphics[width=\linewidth]{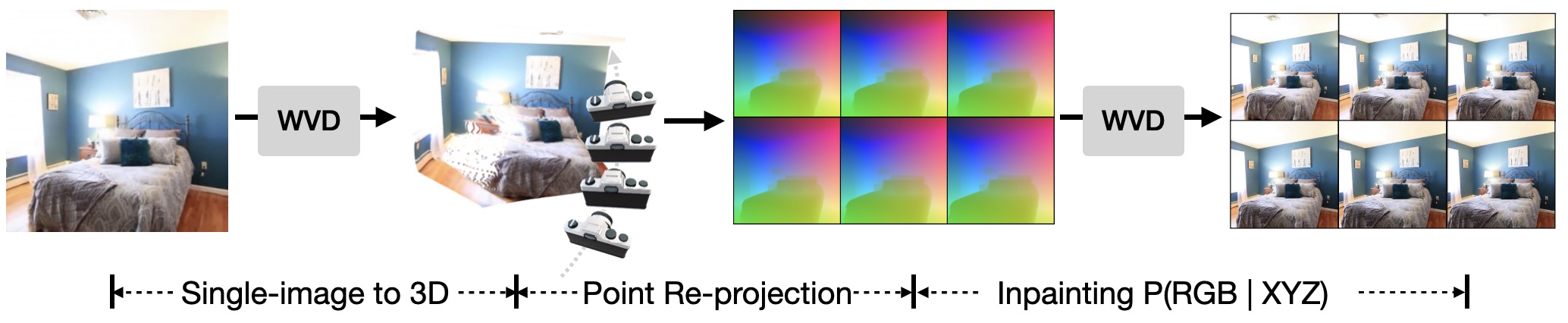}
    \caption{\textbf{Illustration of camera-controlled multi-view generation pipeline.} We first use \NAME to infer the geometry from the input image, and then project it to obtain XYZ images for novel views. Next, we employ an inpainting strategy to sample RGB images. }
    \label{fig:cam-ctrl-pipeline}
\end{figure*}
\subsection{Single-image to 3D Tasks}
Given its training methodology, \Model can be directly applied to various single-image tasks, including monocular depth estimation (as described in \cref{eq.post}), novel view synthesis, and 3D reconstruction. Notably, unlike traditional monocular depth estimation approaches that are typically supervised to infer depth from single-image inputs, our approach estimates depth through a generative process. By jointly sampling consistent surrounding views from the learnt data distribution, \Model produces depth predictions that are more 3D-grounded and consistent with the global scene geometry.

\subsection{Multi-view Stereo Tasks}
\label{sec.mvs}
Since \Model learns the distribution of videos, it can also be applied to multi-view tasks with a collection of \textbf{unposed} RGB images provided. In this setup, the model predicts only the {\XYZ} images through a diffusion process, following a procedure akin to “in-painting”~\cite{Shih20203DInpainting}. At each diffusion step, the model’s RGB predictions are replaced with the observed RGB values, ensuring consistency with the given inputs while generating the missing {\XYZ} components. Consistent with the findings in \cite{gu2024control3diff}, our early experiments revealed that incorporating additional Langevin correction steps~\cite{song2020score} significantly enhances the quality and stability of the in-painting process.
With the additional post-optimization steps (\cref{eq.post}), \Model not only reconstructs 3D geometry but also enables consistent multi-view video depth estimation. This capability makes it highly valuable for applications that require accurate 3D scene interpretation from diverse viewing angles.

\subsection{Controllable Generation Tasks}
\label{sec.control-gen}
\Model also supports controllable video generation by leveraging the {\XYZ} information in reverse mode. Similar to existing multi-view diffusion models, \Model is trained in a camera-agnostic manner, learning the underlying distribution of camera trajectories implicitly without requiring explicit conditional guidance. However, at inference time, the model can be adapted to enable video generation with camera control through point re-projection.
As illustrated in \cref{fig:cam-ctrl-pipeline}, the pipeline involves the following steps:

\begin{enumerate}
    \item \textbf{Single-image to 3D:} We first estimate the points of the input using standard \Model diffusion inference.
    \item \textbf{Point Re-projection:} The synthesized point clouds are projected onto the target camera poses, producing partial {\XYZ} images with corresponding projection masks, as described in \cref{eq.raster};
    \item \textbf{RGB \& {\XYZ} In-painting:} Finally, \Model regenerates the RGB images jointly with the projected {\XYZ} images through an in-painting process.
\end{enumerate}

Unlike the scenario in \cref{sec.mvs}, the projected {\XYZ} images in this case are typically incomplete, requiring the model to in-paint both the RGB and missing {\XYZ} components during the diffusion process.
In addition, the above point-guidance process enables us to maintain the union of synthesized points as a “spatial memory,” where new video frames are guided by the projected points. This approach allows for the progressive generation of long video sequences while enforcing explicit consistency constraints, ensuring coherence across frames.

\section{Experimental Results}

\begin{figure*}[t]
    \centering
    \includegraphics[width=0.97\textwidth]{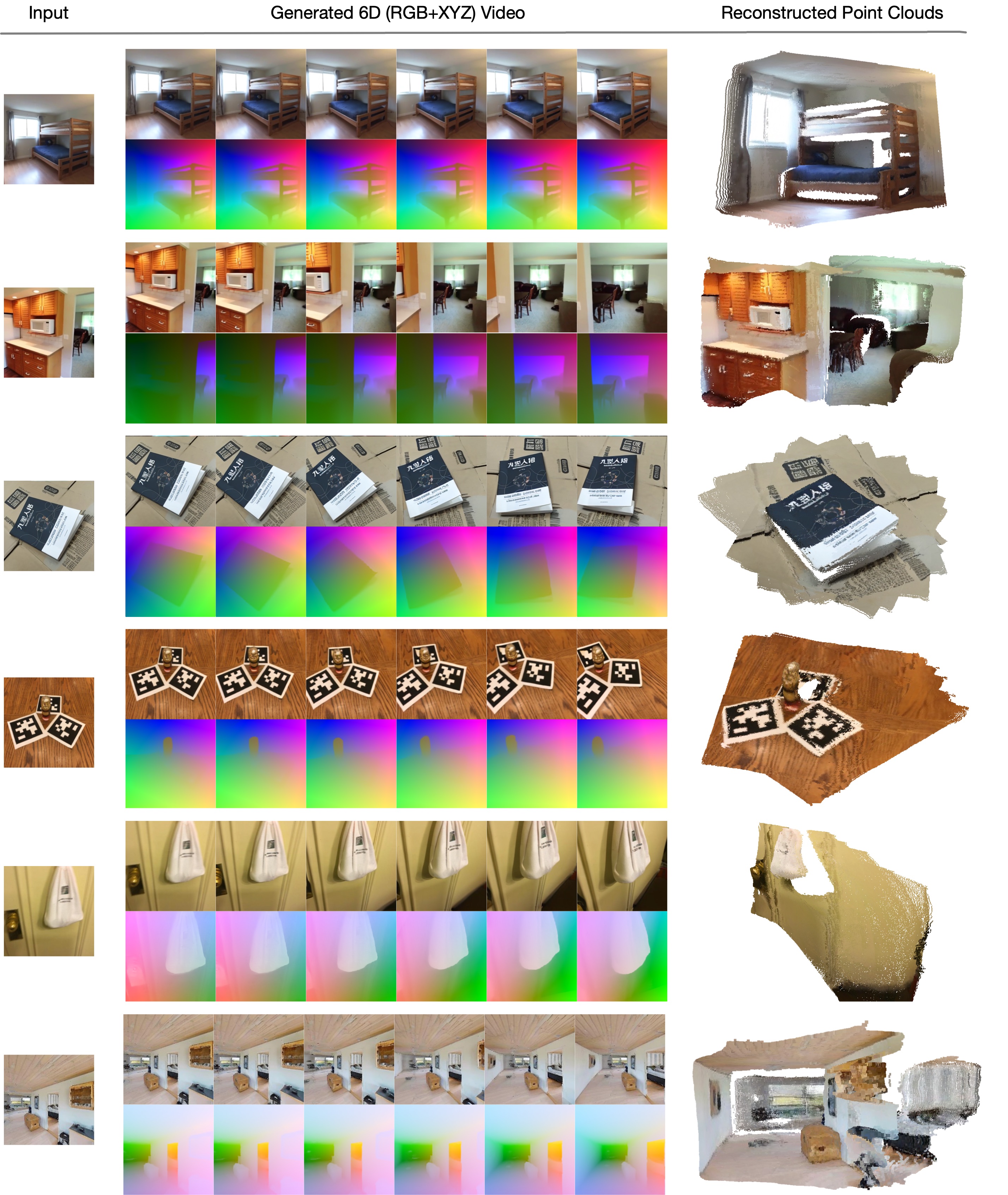}
   \caption{\textbf{Synthesized Multi-view RGB and {\XYZ} Images by \NAME, and associated reconstructed point clouds.} Input images are randomly sampled across the validation set.}
    \label{fig:single-image-to-3D}
\end{figure*}

\subsection{Settings}

\paragraph{Datasets.} We train our model \Model on a mixture of datasets: RealEstate10K~\cite{zhou2018stereo}, ScanNet~\cite{dai2017scannet}, MVImgNet~\cite{yu2023mvimgnet}, CO3D~\cite{reizenstein21co3d}, and Habitat~\cite{habitat19iccv}. These datasets cover a broad range, from object-centric to scene-centric distributions. For RealEstate10K, MVImgNet, and CO3D, we use DUSt3R~\cite{wang2024DUSt3R} to generate pseudo-ground-truth point clouds. ScanNet offers ground-truth depth maps that contain holes, which we fill using NeRF with depth regularization~\cite{kangle2021dsnerf}. For Habitat, we directly utilize the rendered ground-truth point cloud.
All images are center-cropped and resized to $256 \times 256$ resolution.

\vspace{-10pt}\paragraph{Implementation details.} Our Diffusion Transformer has 2 billion parameters and is implemented with rotary positional embedding~\cite{su2024roformer} and RMSNorm~\cite{zhang2019root}. A detailed model card is available in the Supplementary materials. As for training, we employ a learning rate of $3\times10^{-4}$ using the AdamW optimizer, with the momentum parameter setting to $\beta=(0.99, 0.95)$. We train the model for 1 million steps, with an effective batch size of 128. The training takes approximately two weeks over 64 A100 GPUs. 
\begin{figure*}[t]
    \centering
    \includegraphics[width=0.98\linewidth]{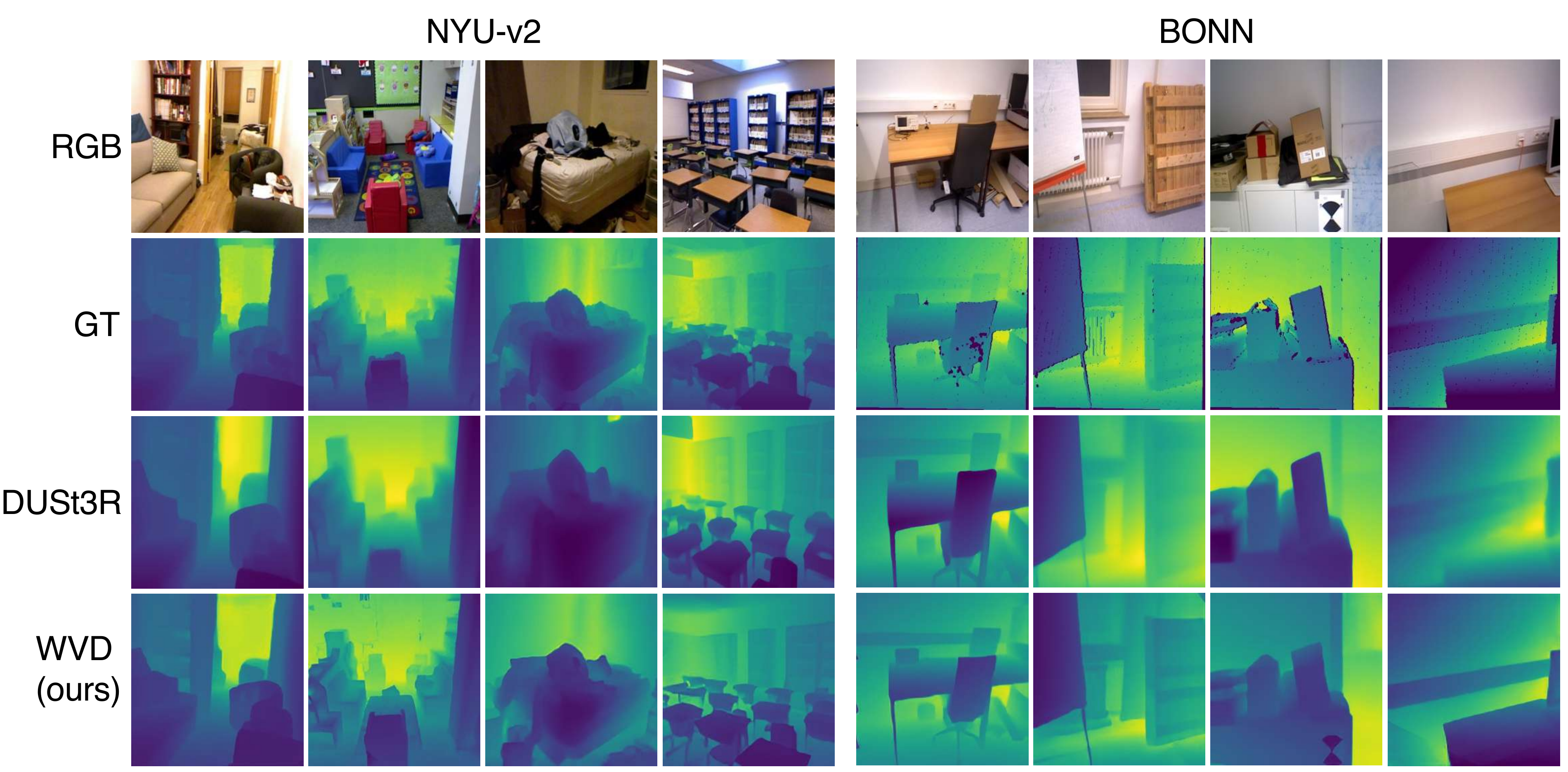}
   \caption{\textbf{Monocular depth estimation on NYU-v2~\cite{Silberman2012} and BONN~\cite{palazzolo2019refusion} benchmarks.} We present RGB input images, ground-truth depth maps, and the predicted depth maps from DUSt3R (512 resolution) and WVD, respectively.}
    \label{fig:mono-dep}
\end{figure*} 

\begin{figure*}[t]
    \centering
    \includegraphics[width=0.96\linewidth]{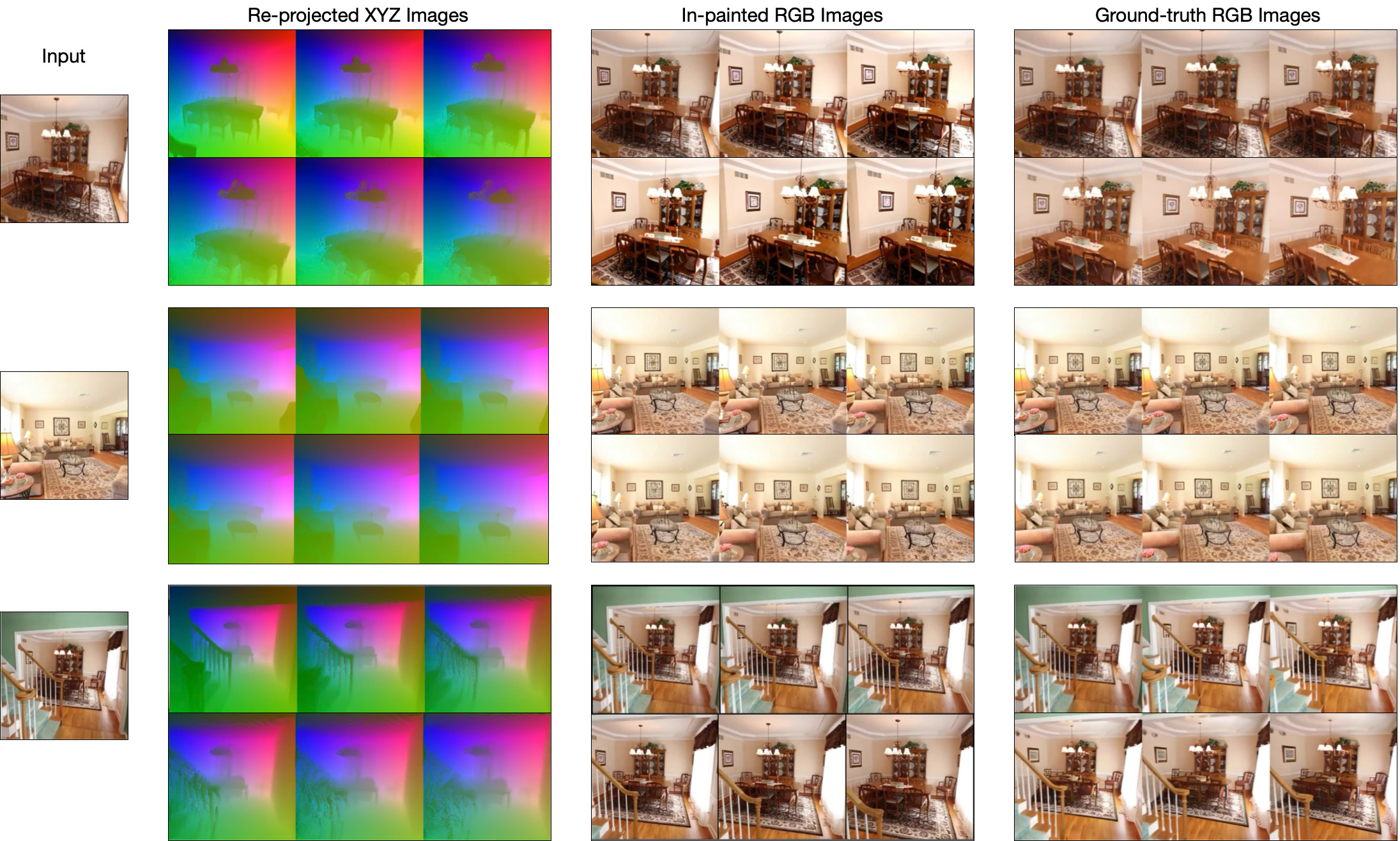}
   \caption{\textbf{Camera-controlled video generation.} By re-projecting XYZ images and using them as conditions, our method can control the camera movements in the synthesized videos, effectively replicating the trajectories of the real videos.}
    \label{fig:cam-ctrl-result}
\end{figure*}

\subsection{Main Results}

\begin{table}[t]
\centering
\caption{Quantitative comparisons for single image to 3D. }
\label{img-to-3D}
\setlength{\tabcolsep}{12pt}
\begin{tabular}{@{}lccc@{}}
\toprule
Method     & FID$\downarrow$   & KPM$\uparrow$  & FC$\uparrow$  \\ \midrule
CameraCtrl~\cite{he2024cameractrl} & \textbf{12.1} & 88.6 & 94.0  \\
MotionCtrl~\cite{wang2024motionctrl} & 12.9 & 68.6 & 94.6 \\ \midrule
\Model       & 15.8   & \textbf{95.8} & \textbf{95.4} \\ 
\Model w/o XYZ & 18.3  &  72.3 & 95.0  \\
\bottomrule
\end{tabular}
\end{table}

\paragraph{Single Image to 3D.} \cref{fig:single-image-to-3D} illustrates the synthesized RGB and {\XYZ} frames conditioned on a single RGB frame. Our method effectively generates multi-view consistent frames with remarkable detail across a diverse range of visual distributions. Furthermore, we visualize the 3D scenes by projecting RGB pixels into 3D space using the corresponding coordinates from the XYZ frames. The resulting 3D point cloud exhibits realistic appearance and geometry, showcasing our ability to create 3D scenes from a single image.

For quantitative comparison with baselines, we choose the following metrics to measure the quality of synthesized frames: (1) Frechet Inception Distance (FID)~\cite{heusel2017gans} measures the per-frame appearance quality; (2) Key Points Matching (KPM)~\cite{wang2024driving} assesses multi-view consistency by averaging the number of matching key points identified by a pretrained matching model~\cite{sun2021loftr}. We use the numbers obtained from the ground truth videos as a baseline and report the percentage of each method; (3) Frame Consistency (FC)~\cite{huang2024vbench} assesses a video based on the similarity of the CLIP image features among its frames. We choose CameraCtrl~\cite{he2024cameractrl} and MotionCtrl~\cite{wang2024motionctrl} as baselines, and show the results in \cref{fig:single-image-to-3D}. 
Our method achieves comparable performance to CameraCtrl and MotionCtrl in terms of frame appearance. It consistently outperforms both baselines in multi-view consistency, as measured by KPM and FC, highlighting the advantages of jointly modeling XYZ images alongside RGB images.

\vspace{-10pt}\paragraph{Monocular depth estimation.} As shown above, our method can generate multi-view RGB and {\XYZ} images with single RGB frame. By converting the {\XYZ}-encoded point clouds into dense depth maps, we enable monocular depth estimation. We evaluated our approach against other zero-shot methods on the NYU-v2~\cite{Silberman2012} and BONN~\cite{palazzolo2019refusion} benchmarks for monocular depth estimation. We visualize the result in \cref{fig:mono-dep}. Although our method is never trained over any depth prediction benchmark, our model can make precise prediction given monocular image as input. Quantitative comparison with baselines is presented in \cref{table:nyu-v2}. On BONN, our method outperforms all baseline models. While DUSt3R trained at $512\times 512$ achieves the best performance on NYU-v2, our model, despite being trained at a lower resolution ($256\times 256$), surpasses all other baseline methods.

\begin{table}[t]
\centering
\caption{Monocular depth estimation performance on NYU-v2~\cite{Silberman2012} and BONN~\cite{palazzolo2019refusion}.
*DUSt3R-512 was trained with higher resolution than ours.
}
\label{table:nyu-v2}
\begin{tabular}{lcccc}
\toprule
\multirow{2}{*}{Methods} & \multicolumn{2}{c}{NYU-v2~\cite{Silberman2012}} & \multicolumn{2}{c}{BONN~\cite{palazzolo2019refusion}} \\ \cline{2-5}
 & Rel $\downarrow$ & \textbf{$\delta_{1.25}$} $\uparrow$ & Rel $\downarrow$ & \textbf{$\delta_{1.25}$} $\uparrow$ \\
\midrule
RobustMIX~\cite{ranftl2020towards} & 11.8 & 90.5 & - & - \\  
SlowTv~\cite{spencer2023kick} & 11.6 & 87.2 & - & - \\  
DUSt3R-224~\cite{wang2024DUSt3R} & 10.3 & 88.9 & 11.1 & 89.1 \\
DUSt3R-512~\cite{wang2024DUSt3R}$^*$ & \underline{6.5} & \underline{94.1} & 8.1 & 93.6 \\  
\midrule
\Model & \textbf{9.7} & \textbf{90.8} & \textbf{7.0} &  \textbf{96.4} \\  
\bottomrule
\end{tabular}
\end{table}

\vspace{-10pt}\paragraph{Video depth estimation.}
As discussed in \cref{sec.mvs}, our model can estimate the conditional distribution of P({\XYZ}~$|$~RGB) using an in-painting strategy. This allows us to adapt our model for estimating 3D geometry based on a set of unposed RGB images. We can sample point clouds from P({\XYZ}~$|$~RGB), and subsequently convert these point clouds into dense depth maps through post-optimization, effectively repurposing our method as a video depth estimator. We benchmark this capability and present the performance in \cref{tab:video-depth}. The results demonstrate that our method performs on par with state-of-the-art approaches..

\begin{table}[t]
\centering
\caption{Video depth estimation performance on ScanNet++.}
\label{tab:video-depth}
\setlength{\tabcolsep}{9pt}
\begin{tabular}{lcc}
\midrule
\multirow{2}{*}{Method} &  \multicolumn{2}{c}{ScanNet++}  \\  \cline{2-3}
& AbsRel $\downarrow$ & $\delta_{1.03} \uparrow$ \\
\midrule
COLMAP~\cite{schoenberger2016mvs, schoenberger2016mvs} &  14.6 & 34.2 \\ \midrule
Vis-MVSSNet~\cite{zhang2023vis}  & 8.9 & 33.5 \\
MVS2D~\cite{yang2022mvs2d}  & 27.2 & 5.3 \\ \midrule
DeMon~\cite{ummenhofer2017demon} & 75.0 & 0.0 \\
MVSNet~\cite{yao2018mvsnet} & 65.2 & 28.5 \\
Robust MVD~\cite{schroppel2022benchmark} & 7.4 & 38.4 \\ \midrule
DeepV2D~\cite{teed2018deepv2d}  & \textbf{4.4} & 54.8 \\
DUSt3R-224~\cite{wang2024DUSt3R} & 5.9 & 50.8 \\
DUSt3R-512~\cite{wang2024DUSt3R}$^*$ & 4.9 & \underline{60.2} \\ 
\midrule
\Model & 5.0 & \textbf{57.2}  \\ 
\midrule
\end{tabular}
\end{table}

\vspace{-10pt}\paragraph{Camera-controlled Video Generation.}
As mentioned in \cref{sec.control-gen}, our model enables camera-controlled video generation. This is accomplished by first estimating 3D geometry from a single input image, which is then used to guide the generation of novel views.

We demonstrate this process in \cref{fig:cam-ctrl-result}, showcasing the ground-truth videos, the corresponding projected XYZ images, and the videos generated by our method. The synthesized videos can mimic the camera motion observed in the real videos, highlighting the effectiveness of our approach for camera-controlled video synthesis.

\vspace{-10pt}\paragraph{Ablation of jointly predicting {\XYZ}   together with RGB frames.}
In \cref{img-to-3D}, we assess the necessity of learning  {\XYZ} frames by training an RGB-only model. Without learning the {\XYZ} frames, both image quality and multi-view consistency declines. This demonstrates that jointly learning the {\XYZ} frames offers explicit 3D supervision, which enhances multi-view synthesis.
\section{Discussions and Future Work}

We introduce \NAME, a DiT framework that jointly models the distribution of multi-view RGB and XYZ images, enabling direct 3D scene generation without the need for post-processing. Additionally, \NAME can be adapted for various downstream tasks (e.g., monocular depth estimation, camera pose estimation) through a flexible inference strategy.

While \NAME demonstrates the potential to serve as a 3D foundation model, the framework itself is not limited by modality. Future work could explore incorporating different modalities rather than 3D XYZ images (e.g., optical flow, splatter images) within our framework to support an even broader range of tasks.

\vspace{-10pt}\paragraph{Limitations.} Our model currently has the following limitations: (1) We have only trained on static datasets, restricting its application to static scenes. Extending this work to dynamic 4D datasets and jointly learning motion-related representations, such as optical flow, would be a valuable direction for future research. (2) Our model does not incorporate confidence maps, making it challenging to handle unbounded or outdoor scenes. Jointly modeling {\XYZ} with confidence could improve performance in such scenarios.
{
    \small
    \bibliographystyle{ieeenat_fullname}
    \bibliography{main}
}
\newpage
\appendix
\newcommand{\AppendixPrefix}{A}
\renewcommand{\thefigure}{\AppendixPrefix\arabic{figure}}
\setcounter{figure}{0}
\renewcommand{\thetable}{\AppendixPrefix\arabic{table}} 
\setcounter{table}{0}
\renewcommand{\theequation}{\AppendixPrefix\arabic{equation}} 
\setcounter{equation}{0}

\section*{Appendix}

\section{Implementation Details}

We begin by training a text-to-video Latent Diffusion Model (LDM) on web-scale datasets, which serves as the initialization for \NAME. For this, we utilize the Variational Autoencoder (VAE) from SDXL~\cite{podell2023sdxl}. We implement 3D self-attention to capture the spatial and temporal relationships between image patches. In addition to the timestep, we incorporate a binary mask to specify whether the frames are RGB or XYZ. This conditioning information is integrated through cross-attention. We remove the text embedding from the original model. 

\paragraph{Classifier-free guidance.} During training, we randomly select either a single RGB frame or a single XYZ frame for conditioning. With classifire-free-guidance (CFG) implemented, we randomly drop conditional images with a probability of 0.1 during training. The denoising score can be written as:
\begin{equation}
    \bm{\epsilon} = (1+w)(k\bm{\epsilon}^{\textrm{RGB}} + (1-k)\bm{\epsilon}^{\textrm{XYZ}}) - w\bm{\epsilon}^\textrm{Uncond},
\end{equation}
where $\bm{\epsilon}^{\textrm{RGB}}$, $\bm{\epsilon}^{\textrm{XYZ}}$, and $\bm{\epsilon}^{\textrm{Uncond}}$ represent the estimated scores with RGB conditioning, XYZ conditioning, and unconditional score, respectively. $w$ is the guidance strength, and $k$ balances the guidance between RGB and XYZ. In most cases, we select $k=1$ to condition solely on RGB frames. For camera-controlled video generation, we use 
$k=0.5$.

\section{Camera-controlled Video Generation.}

\begin{figure*}[t]
    \centering
    \includegraphics[width=0.98\linewidth]{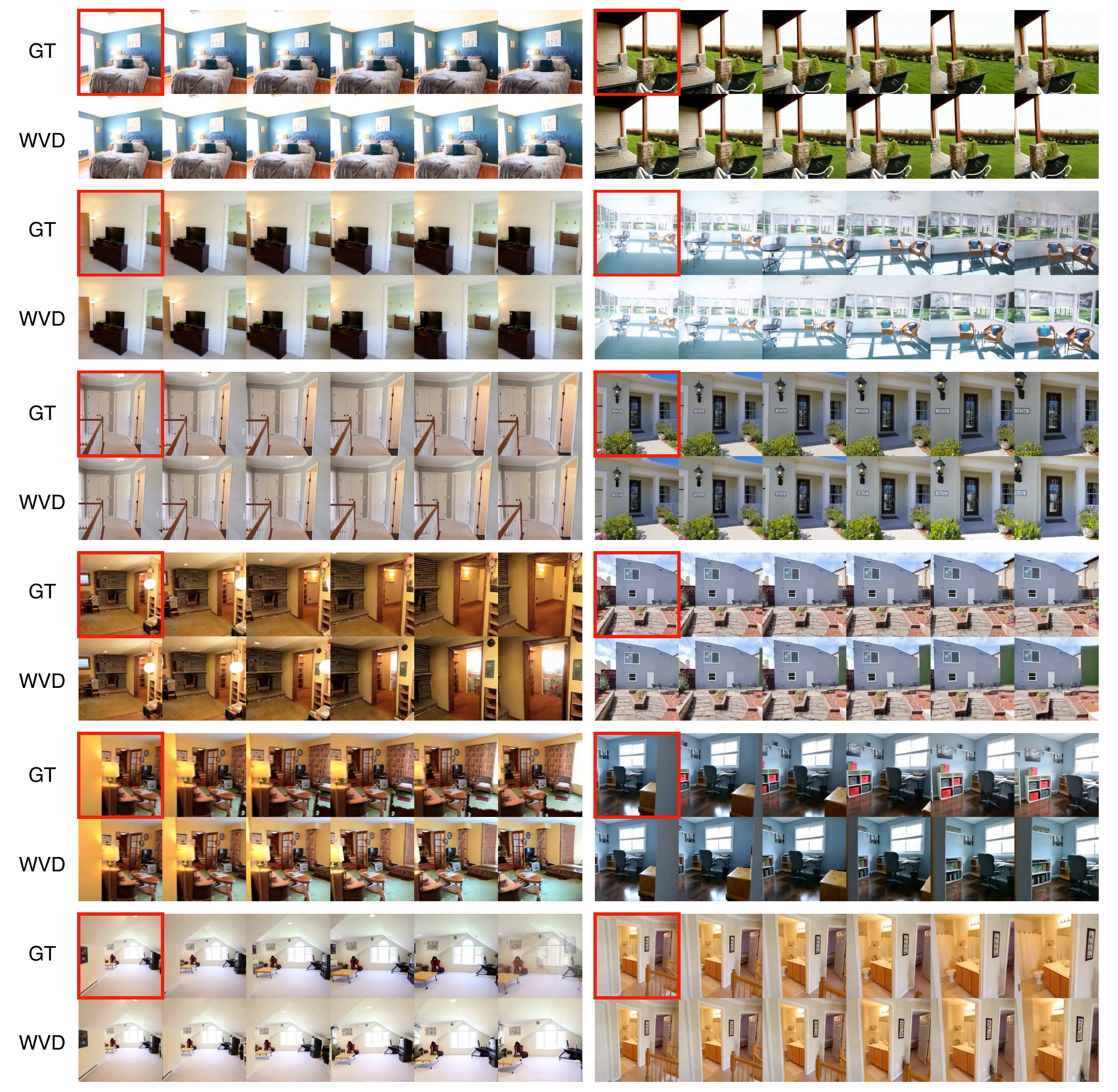}
   \caption{\textbf{Camera-controlled video generation.} For each sample, the first row shows the ground-truth video sequence, and the second row shows the synthesized frames which re-produce the camera trajectory. The conditioned frame is marked with a red box.}
    \label{fig:supp-cam-ctrl-result}
\end{figure*}

We present additional samples for camera-controlled video generation using the test set from RealEstate10K~\cite{Zhou2018StereoImages} in \cref{fig:supp-cam-ctrl-result}. As shown, the synthesized videos closely replicate the camera motion observed in the ground-truth videos, highlighting our model's camera-control capability. It's important to note that camera information was not utilized during training.





\begin{figure*}[t]
    \centering
    \includegraphics[width=\linewidth]{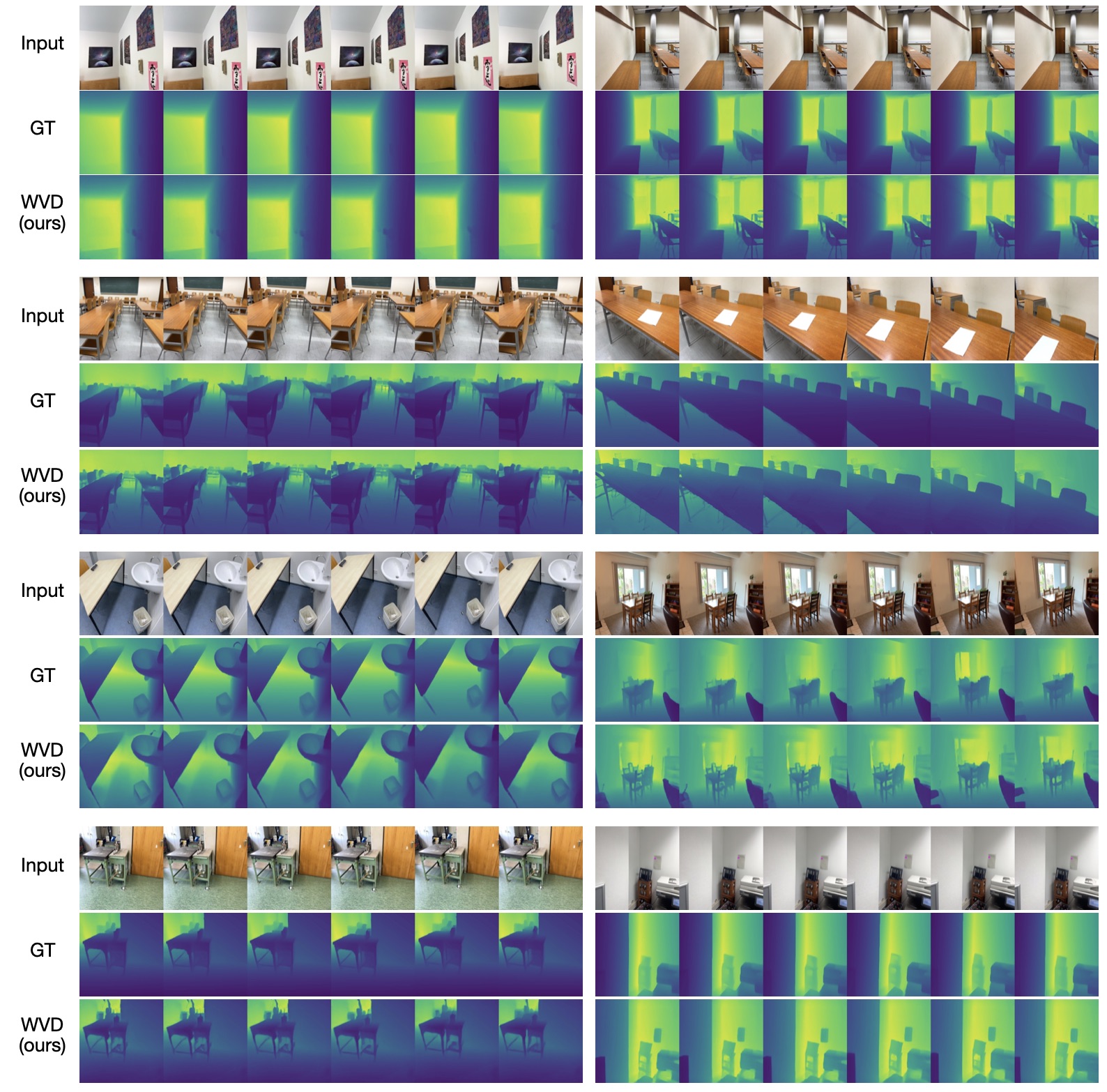}
   \caption{\textbf{Multi-view depth estimation on ScanNet++~\cite{yeshwanth2023scannet++}.} For each sample, the first row presents the input video sequence, the second row shows the ground-truth depth maps. The third row shows the depth maps synthesized by our method.}
    \label{fig:supp-mv-depth}
\end{figure*} 

\begin{figure*}[t]
    \centering
    \includegraphics[width=\linewidth]{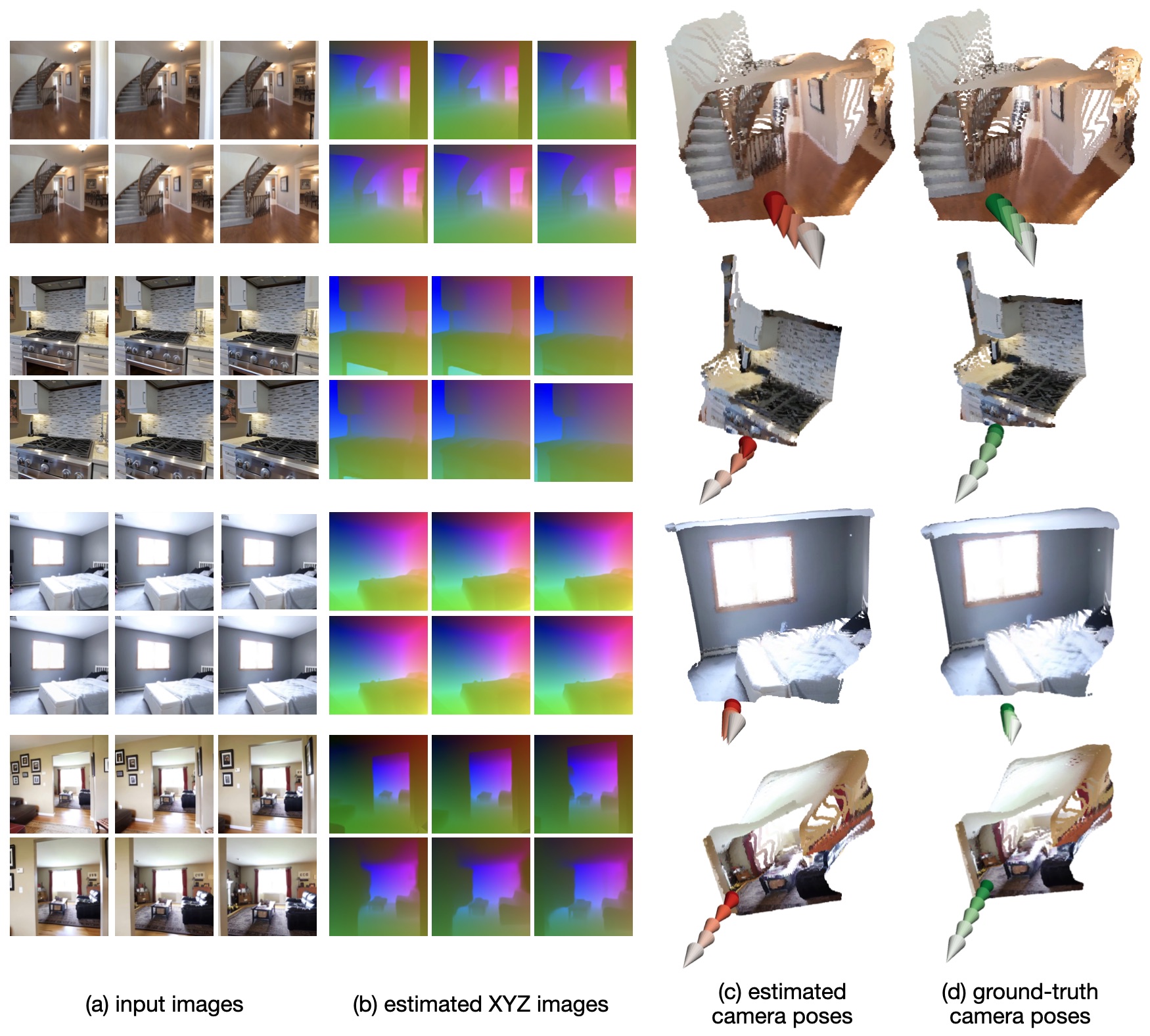}
   \caption{\textbf{Camera estimation.} Column (a) shows input unposed images. Column (b) shows estimated XYZ images by our method. Column (c) shows the estimated camera poses from the XYZ images, while column (d) provides the ground-truth camera poses.}
    \label{fig:supp-cam-est}
\end{figure*} 

\begin{figure*}[t]
    \centering
    \includegraphics[width=\linewidth]{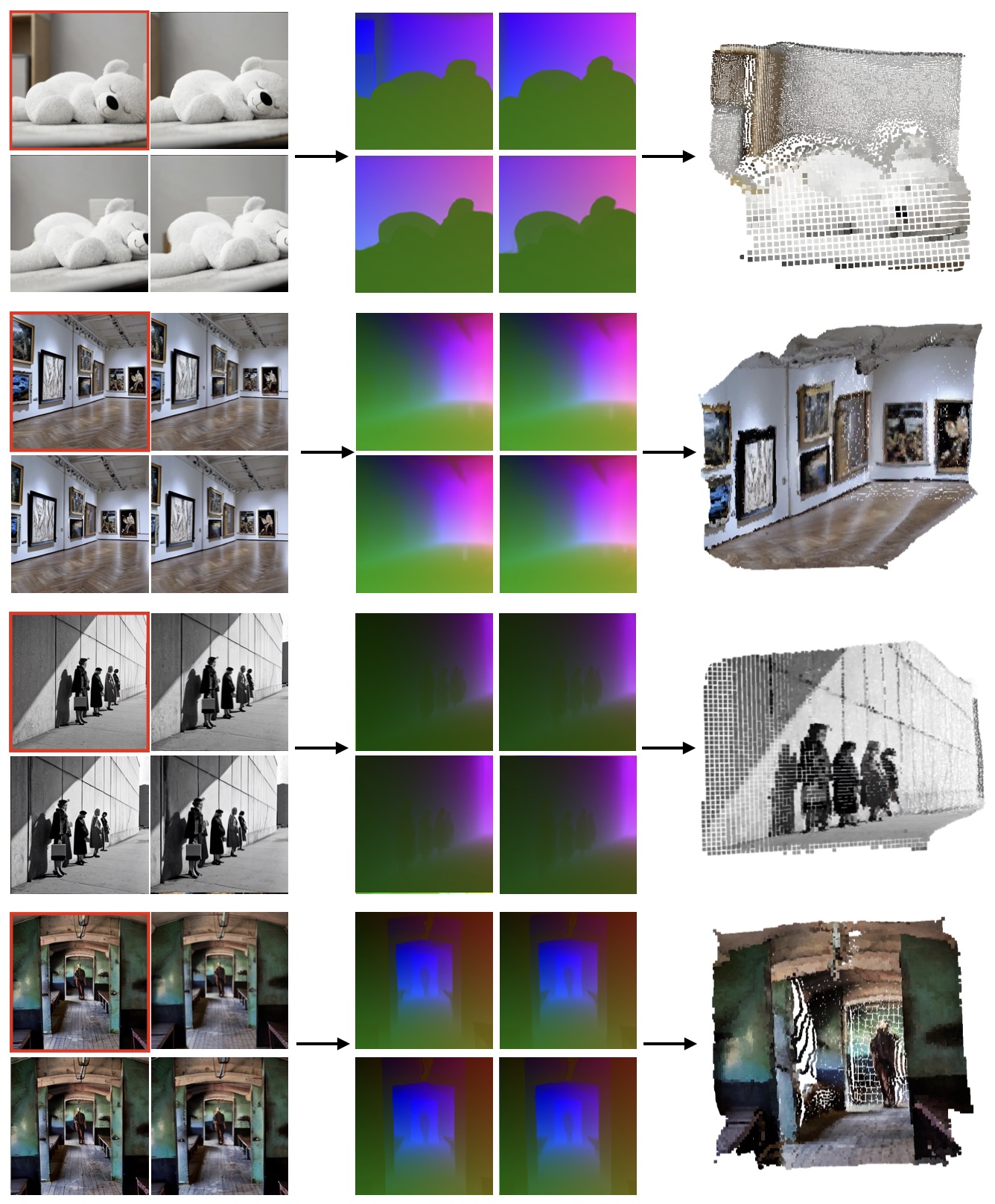}
   \caption{\textbf{In-the-wild samples.} We evaluate our model on in-the-wild samples to demonstrate its generalizability. The conditioned image is highlighted with a red box.}
    \label{fig:supp-ood}
\end{figure*} 

\section{Multi-view Depth Estimation}

We present samples over ScanNet++~\cite{yeshwanth2023scannet++} for multi-view depth estimation in \cref{fig:supp-mv-depth}. As demonstrated, our method can accurately estimate depth maps with video input.

\section{Camera Estimation}

As outlined in the main paper, our method can predict the corresponding XYZ frames through inpainting when provided with ground-truth RGB frames. In addition to video depth estimation, these XYZ frames can also be utilized for camera estimation tasks. Specifically, we can easily perform a Perspective-n-Point (PnP) algorithm to extract the camera poses from point clouds represented by the XYZ frames. We use gradient optimization over re-projection loss as specified in the main paper.
\cref{fig:supp-cam-est} presents results from the test set of RealEstate10K~\cite{Zhou2018StereoImages}. Our method not only predicts accurate 3D geometry from unposed images but also estimates precise camera trajectories. The estimated camera poses are in close agreement with the ground truth.

\section{In-the-wild Samples}
We also evaluate our model on in-the-wild samples to assess its generalizability. As illustrated in \cref{fig:supp-ood}, our model successfully generalizes to out-of-domain images, such as those generated by AIGC algorithms. It can produce novel view images, accurately estimate XYZ images, and reconstruct 3D scenes from a single image. This demonstrates that our method exhibits strong generalizability after training on a diverse mixture of datasets.

\end{document}